\documentclass[10pt,twocolumn,letterpaper]{article}
\usepackage[pagenumbers]{cvpr} 

\usepackage{graphicx,epstopdf}
\usepackage{amsmath}
\usepackage{amssymb}
\usepackage{booktabs}
\usepackage{diffcoeff}  
\usepackage{multirow}
\usepackage[colorlinks=true,linkcolor=blue,urlcolor=blue,citecolor=blue,anchorcolor=red]{hyperref}
\DeclareMathAlphabet{\mathpzc}{OT1}{pzc}{m}{it}
\usepackage{epstopdf}

\usepackage{xcolor}

\usepackage[capitalize]{cleveref}
\crefname{section}{Sec.}{Secs.}
\Crefname{section}{Section}{Sections}
\Crefname{table}{Table}{Tables}
\crefname{table}{Tab.}{Tabs.}

\DeclareMathOperator*{\argmin}{argmin} 

\def\cvprPaperID{9686} 
\def\confName{CVPR}
\def\confYear{2022}

\begin{document}

\title{\LARGE DiffPoseNet: Direct Differentiable Camera Pose Estimation}

\author{Chethan M. Parameshwara, Gokul Hari, Cornelia Ferm{\"u}ller, Nitin J. Sanket, Yiannis Aloimonos \\ \\
Perception and Robotics Group,
University of Maryland, College Park, MD, USA \\
{\tt\small \{cmparam9,hgokul,fer,nitin,yiannis\}@umiacs.umd.edu}
}
\maketitle

\begin{abstract}


Current deep neural network approaches for camera pose estimation rely on scene structure for 3D motion estimation, but  this decreases the robustness and thereby makes cross-dataset generalization difficult. In contrast, classical  approaches to structure from motion estimate 3D motion utilizing optical flow and then compute depth. Their accuracy, however, depends strongly on the quality of the optical flow. To avoid this issue, direct methods have been proposed, which  separate 3D motion from depth estimation, but compute 3D motion using only image gradients in the form of normal flow. In  this  paper, we introduce a network \textit{NFlowNet}, for normal flow estimation which is used to enforce robust and direct constraints. In particular, normal flow is used to estimate relative camera pose based on the cheirality (depth positivity) constraint. We achieve this by formulating the optimization problem as a differentiable cheirality layer, which allows for end-to-end learning of camera pose. We perform extensive qualitative and quantitative evaluation of the proposed DiffPoseNet's sensitivity to noise and its generalization across datasets. We compare our approach to existing state-of-the-art methods on KITTI, TartanAir, and TUM-RGBD datasets.

\end{abstract}

\section{Introduction}
\label{sec:intro}


The ability to localize is imperative for applications in mobile robotics, and solutions based on vision are often the preferred choice because of size, weight, power constraints and the availability of robust localization methods. Many 
mathematical frameworks and deep learning approaches have been developed for the problem of visual localization \cite{scaramuzza2011visual,davison2007monoslam} under the umbrella of Visual Odometry (VO) or Simultaneous Localization and Mapping (SLAM). However, their performance is subpar for commonly encountered challenging conditions in-the-wild that involve changing 
lighting, scenes with textureless regions, and dynamic objects.

\begin{figure*}[t!]
\begin{center}
    \includegraphics[width=\textwidth]{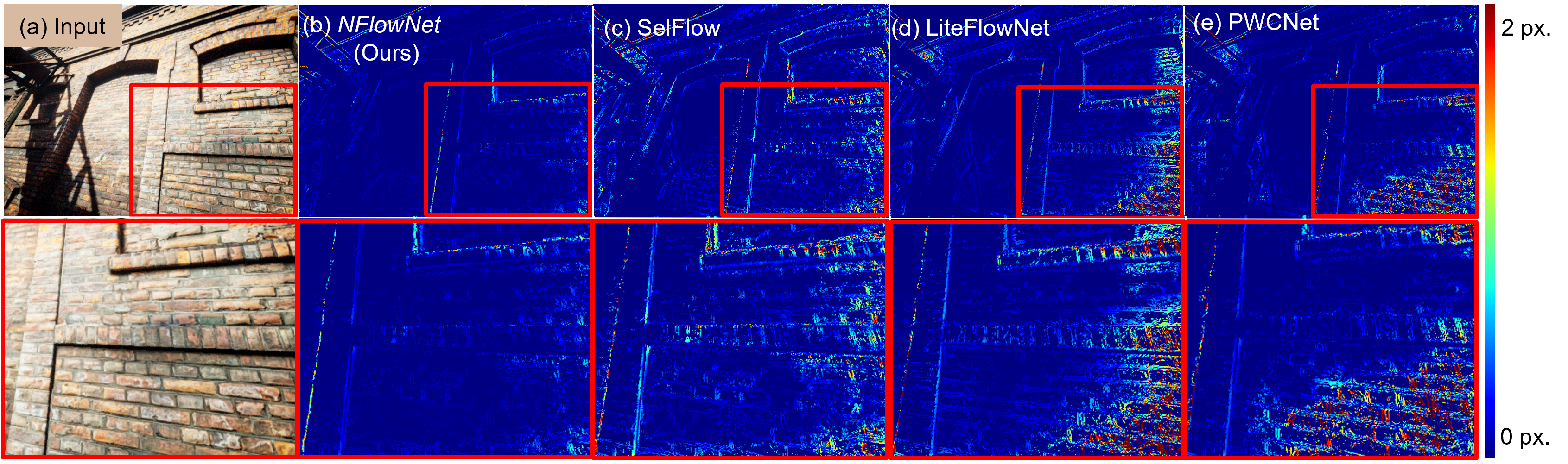}
\end{center}
   \caption{Top Row: Endpoint error map of \textit{NFlowNet} compared to three different optical flow approaches (SelFlow \cite{selflow}, LiteFlowNet \cite{liteflownet}, and PWC-Net \cite{pwc-net}). Bottom Row: enlarged endpoint error map of the region highlighted in red in the top row. The optical flow is due to the camera undergoing translation parallel to the wall. It has large errors in regions of non-uniform gradient directions. Notably, on the bricks of rectangular shape, there are many more (vertical) gradients due to the horizontal edges than (horizontal) gradients due to vertical edges, and this causes erroneous flow estimation. Similarly, on the edges of the niches, where there is only one gradient direction, there is error. \textit{All the images in this paper are best viewed in color on a computer screen at a zoom of 200\%.}} 
\label{fig:nfvsof}
\end{figure*}

Classical approaches \cite{black1996robust, brox2004high, ji2007better, brox2010large} for localization rely either on sparse feature correspondences between images or on the computation of dense motion fields (optical flow). One of the difficulties in optical flow estimation is bias due to noise \cite{fermuller2001statistics,fermuller2000ouchi}. For example, if in a patch there are more gradients in one direction than another, their estimated optical flow will be biased towards the dominant direction. Even though over the past decade many learning-based approaches have proposed to improve optical flow estimation, this behavior still persists in optical flow approaches. This is demonstrated in Fig.~\ref{fig:nfvsof}, which shows the errors produced by the normal flow algorithm presented in this paper in comparison to three optical flow algorithms from the literature. As can be seen, all flow algorithms have large errors in regions of non-uniform gradient distributions.

To this end, the pioneers of the field remarked on the observation that the projection of optical flow on the image gradient direction is resilient to the bias and can be computed robustly. This projection is called the \textit{normal flow}. Over the past few decades, a number of methods have been proposed that use the spatial gradient directly for 3D motion recovery. These methods are commonly called direct methods. In principle, such methods are robust and computationally cheaper than flow-based feature based approaches as they use the image brightness directly. However, despite  the advantages of the normal flow formulation, the computational methods to estimate normal flow have not been robust enough to allow deployment in the wild. Thus, optical flow has been the go-to representation for ego-motion estimation, supported in recent years by the high accuracy and speed of deep learning algorithms \cite{ilg2017flownet,jonschkowski2020matters,ranjan2017optical}. To improve the robustness of camera pose estimation (ego-motion), we propose the first normal flow network \textit{NFlowNet}. 

Further, to estimate pose independent of scene structure from normal flow, direct approaches utilize minimal constraints.
When optical flow or correspondence are  available, pose is estimated using the depth-independent epipolar constraint. However different from these 2D measurements, normal flow is 1D and thus depth cannot be eliminated from the equations relating it to scene geometry and 3D motion. Not making assumptions about the scene structure,
the only constraint that can be imposed on the scene, is the  depth positivity \cite{negahdaripour1986direct} or cheirality constraint \cite{Wrobel2001MultipleVG}. Cheirality states that the scene has to be in front of the camera for it to be imaged, and thus the depth has to be positive. This constraint when enforced on normal flow can be utilized to estimate camera pose without making assumptions on scene depth or shape. Since the cheirality condition is an inequality constraint and hence is not differentiable, until recently it was not possible to employ it in a deep learning pipeline. To this end, we utilize the differentiable programming paradigm \cite{amos2017optnet}  implemented with the implicit differentiation \cite{gould2021deep} framework to reformulate the cheirality optimization into a differentiable layer and hence train our pose network in an end-to-end fashion.

In this work, we design a novel normal flow network \textit{NFlowNet} and  couple it with a differentiable cheirality layer for robust pose estimation. Our contributions (in the order for ease of understanding) can be summarized as follows:
\begin{itemize}
    \item We introduce a network \textit{NFlowNet} to estimate normal flow. This estimated robust normal flow,  beyond this paper, is useful for  applications requiring computationally efficient solutions for navigation tasks in computer vision and robotics. 
    \item We formulate the estimation of pose from  normal flow using the cheirality (or depth positivity) constraint as a differentiable optimization layer.
    \item 
    Extensive qualitative and quantitative experimental results highlighting the robustness and cross-dataset generalizability of our approach without any fine-tuning and/or re-training.
\end{itemize}

\section{Related Work}
\label{sec:related}

\subsection{Normal Flow and Camera Pose}
Several works have developed direct methods that use normal flow for pose estimation.
The idea is that normal flow can be interpreted as ``the projection of optical flow on the gradient direction." Thus, given a normal flow vector, the optical flow is constrained to a half-plane \cite{aloimonos1994estimating}. If the 3D motion is only due to translation, this constrains the focus of expansion, i.e., the intersection of the translation axis with the image plane \cite{negahdaripour1986direct} to a half plane. 
Based on this concept, \cite{horn1988direct} proposed different algorithms to solve for the case of  translation only,
and \cite{sinclair1994robust} analyzed the method's stability in the presence of small rotations.
Modeling the scene as piece-wise planar, \cite{brodsky1999shape} solved for 3D motion and calibration \cite{brodsky2002self}, and \cite{ji20063d} added constraints for combining multiple flow fields.
Not making depth assumptions, \cite{fermuller1995passive} developed constraints on the sign of normal flow, which geometrically separate the rotational and translational flow components and can be implemented as pattern matching.
Others proposed techniques for separating 3D motion components by searching for lines in the image, where certain 3D motion components cancel out \cite{silva1997robust,yuan2013direct}.
Recently \cite{barranco2021joint} modeled the  cheirality constraint by approximating it with a smooth function, which allowed the  use of modern optimization techniques.
The method first solves for 3D motion from normal flow, and then refines using a regularization defined on depth. Experimental results demonstrate that the proposed pipeline 
outperforms other flow based approaches. 
Inspired by these findings, we follow a similar pipeline, but we develop the constraints within a neural network approach. 


    
    
    
    
    
    
    
    

\subsection{Learning-based Camera Pose Estimation}

Early studies of learning-based camera pose (VO) models \cite{wang2017deepvo} \cite{posenet} \cite{posenet2} were mainly focused on supervised learning approaches modelled as either absolute pose/relative pose regression problems. However, these methods require real-world ground truth poses which is often difficult to obtain. In order to alleviate the need of ground truth data, self-supervised VO was proposed. SfMLearner\cite{zhao2020maskflownet} learns depth and pose simultaneously by minimizing photometric loss between warped and input image. \cite{yin2018geonet} and \cite{zhao2020} extend this idea to joint estimation of pose, depth and optical flow. Learning-based models suffer from generalization issues when tested on images from a new environment. Most of the VO models are trained and tested on the same dataset. Most recently, TartanVO \cite{wang2020tartanvo} addresses generalization issues by  incorporating the camera intrinsics directly into the model and training with a large amount of data. Recent advances in differentiable optimization layers ( or differentiable programming) \cite{amos2017optnet,gould2021deep} has enabled a new generation of generalizable pose learning approaches. \cite{jiang2020joint} embeds an Epipolar geometric constraint into a self-supervised learning framework via bi-level optimization of camera pose and optical flow. BlindPnP \cite{campbell2020solving} embeds geometric model fitting algorithms (PnP algorithm, RANSAC) into implicit differentiation layers. All the above work focus on reconstructing the structure (either through network or optimization) and/or optical flow correspondences for estimating camera motion. In our work, we address robust pose estimation by depending only on structure-less cheirality constraints and normal flow. Finally, we utilize the concepts of best of both-worlds to enable speedup using data prior and novel data generalization from mathematical optimization.


    
    
    






\begin{figure*}

  \setlength{\linewidth}{\textwidth}
  \setlength{\hsize}{\textwidth}
    \begin{center}
          \includegraphics[width=\textwidth]{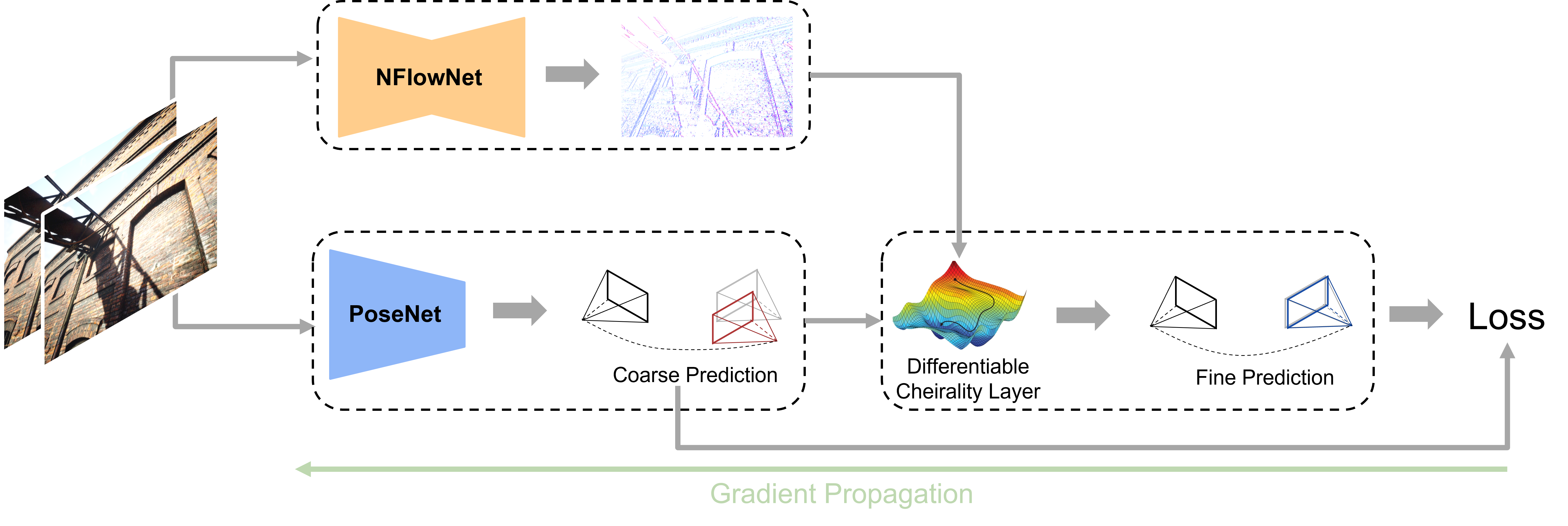}  
    \end{center}

    \caption{Overview of our proposed \textit{DiffPoseNet} framework. Our network starts with a novel normal flow estimation network \textit{NFlowNet} and a first coarse pose estimate. Next, fine pose is estimated using the proposed differentiable cheirality layer.}
    \label{fig:overview}
    
\end{figure*}

\section{Overview of Proposed Approach}

The network architecture is illustrated in Fig.~\ref{fig:overview}. It consists of
\textit{NFlowNet}, a network for estimating normal flow  (Section \ref{NFlowNet}),
which then is used for self-adaptive camera pose estimation  (or odometry estimation) (Section \ref{self-adaptive}).
The camera pose estimation  proceeds in three steps. First we  \textit{initialize} the PoseNet using supervised training with successive images as input (Sec. \ref{subsec:coarse_posenet}). Next, pose is estimated using differentiable optimization by embedding a cheirality constraint defined on normal flow (Sec.~\ref{subsec:cheirality}). This way, simliar to the classic SfM approach, pose is \textit{estimated independent of depth}. In a last step, the pose in PoseNet is refined through a self-supervised refinement loss by using the pose estimates from the \textit{cheirality constraint} to minimize the error in normal flow.

\section{\textbf{\textit{NFlowNet}} for Normal Flow Prediction}

\label{NFlowNet}

    
    

   

The first step in  motion analysis from video is to compute an image motion representation. Most approaches either  detect and track distinct features or compute gradient-based optical flow. 
The latter is estimated by assuming  the intensity $I_{\mathbf{x}}$ (or a function of the intensity) at a point  $\mathbf{x}=\begin{bmatrix} x & y\end{bmatrix}^T$ to remain constant over a short time interval $\delta t$. This  is referred to as the brightness constancy constraint \cite{Wrobel2001MultipleVG}: 
\begin{equation}
\label{brightness1}
I(x,y,t) = I(x+u\delta t, y+v\delta t, t + \delta t)
\end{equation}

Here $u$ and $v$ refers to the image pixel motion. Approximating Eq.(\ref{brightness1}) with a first order Taylor expansion, we obtain Eq.(\ref{brightness2}):
\begin{equation}
\label{brightness2}
\diffp{I}{x} u + \diffp{I}{y}v =- \diffp{I}{t}   \end{equation}

We call the component of the flow along the gradient direction, the \textit{normal flow}.
Denoting the spatial gradients as  $\nabla I = \left(\diffp{I}{x}, \diffp{I}{y} \right)$ and the flow as $\mathbf{{u}}=\begin{bmatrix} u & v\end{bmatrix}^T$, the normal flow vector $\mathbf{\mathfrak{n}}$ (a 2d vector)
is defined as: \begin{equation}
\label{normal_flow}
\mathbf{\mathfrak{n}} = \frac{(\mathbf{u} \cdot \nabla I)}{||\nabla I||^2} \nabla I
\end{equation}

Using the brightness constancy constraint (eq. \ref{brightness1}), the normal  can be computed directly from the spatial and temporal image derivatives, $I_t$, as 
$\mathbf{\mathfrak{n}} = \frac{-I_t}{||\nabla I||^2}\nabla I$.
Since this constraint alone is not sufficient to determine
two-dimensional image motion, additional constraints need to be introduced. Traditionally, variational methods combining multiple global smoothness assumptions, have been the dominant approach for optical flow estimation, and more recently they have been replaced by deep learning algorithms.
However, these methods tend to perform subpar on regions with little features or repeated texture due to the strong reliance on the training dataset, and in boundary regions
due to oversmoothing, especially when the size of the flow varies significantly in the image. Computing normal flow solely on spatio-temporal derivatives, is unreliable and prone to errors. Hence, we propose a novel Normal flow network called \textit{NFlowNet}. 



We use  an encoder-decoder convolutional neural network, which we train in a supervised way. Given an image pair, normal flow describes the pixel motion parallel to the image derivatives. 
To learn normal flow, we utilize the TartanAir dataset. Specifically, we utilize Eq. \ref{normal_flow} to compute  ground-truth normal flow. We train the\textit{NFlowNet}  supervised using the $l_2$ loss between our network predictions $\mathbf{\tilde{n}}$ and ground truth $\mathbf{\hat{n}}$, i.e.,

\begin{equation}
    \label{eqn:nfloss}
    \argmin_{\mathbf{\tilde{\mathfrak{n}}}} \Vert \mathbf{\hat{\mathfrak{n}}} - \mathbf{\tilde{\mathfrak{n}}}\Vert_2
\end{equation}


In the experimental section (Sec. \ref{sec:conclusion}), we show that \textit{NFlowNet} generalizes to the real-world and other datasets without any fine-tuning or re-training.  




\section{Self-Adaptive Pose Estimation from Normal Flow}
\label{self-adaptive}

We use a deep network to regress relative poses, that is, the 3D rigid motion of the camera between subsequent time steps and  denoted as $\mathbf{P}_{t}^{t+1}$. The conversions between absolute poses and relative poses is explained next. 

If the absolute pose at time $t$ is given by $\begin{bmatrix} \mathbf{T}_t & R_t \end{bmatrix}^T$, where $\mathbf{T}_t \in \mathbb{R}^{3 \times 1}$ and $R_t \in SO(3)$. The relative pose between $t$ and $t+1$ under a linear velocity assumption is denoted as $\begin{bmatrix} \mathbf{V} & \mathbf{\Omega} \end{bmatrix}^T$ and is given by

\begin{equation}
    \mathbf{V} = \frac{\mathbf{T}_{t+1}-\mathbf{T}_t}{dt}; \quad \mathbf{\Omega}_\times  = \frac{\text{logm}\left(R^T_tR_{t+1}\right)}{dt}
\end{equation}

Here, $dt$ is the time increment between $t$ and $t+1$, logm is the matrix logarithm operator and $\mathbf{\Omega}_\times$ converts the vector $\mathbf{\Omega}$ into the corresponding skew symmetric matrix

\begin{equation}
\mathbf{\Omega}_\times = \begin{bmatrix} 0 & -\Omega_z & \Omega_y\\ \Omega_z & 0 & -\Omega_x\\ -\Omega_y & \Omega_x & 0\end{bmatrix}
\end{equation}


\subsection{PoseNet for Initializing Pose Estimation}
\label{subsec:coarse_posenet}
In the first stage, we learn coarse relative poses using a CNN+LSTM. The feature representations learned by the CNN layers are passed to the LSTM for sequential modelling. We use  a supervised $l_2$ loss between the ground truth ($\mathbf{\widehat{P}}=\begin{bmatrix} \mathbf{\widehat{V}} & \mathbf{\widehat{\Omega}} \end{bmatrix}^T$) and predicted poses ($\mathbf{\widetilde{P}}=\begin{bmatrix} \mathbf{\widetilde{V}} & \mathbf{\widetilde{\Omega}} \end{bmatrix}^T$). Here, $\mathbf{V}$ and $\mathbf{\Omega}$ represent the translational  and rotational parts of the pose. The orientation is represented in $X-Y-Z$ Euler Angles. Denoting as $\lambda$ a weighting parameter, we solve the following optimization using backpropagation: 

\begin{equation}
    \argmin_{\mathbf{\tilde{P}}} \left( \Vert \mathbf{\widehat{V}} - \mathbf{\widetilde{V}}\Vert_2^2 + \lambda \Vert \mathbf{\widehat{\Omega}} - \mathbf{\widetilde{\Omega}}\Vert_2^2 \right)
\end{equation}

\subsection{Differentiable Cheirality Layer for Fine Pose Estimation}
\label{subsec:cheirality}
To enable self-supervised learning of continuous pose (given a initialization value), we propose to utilize the cheirality constraint or depth positivity constraint, which states that all world points have to be in front of the camera, i.e., have positive depth.
This condition has been classically used in structure from motion problems to disambiguate the physically possible camera poses from the set of computed solutions.
The main reason for utilizing the cheirality  rather than the acclaimed epipolar constaint or scene planarity constraint is due to the minimalism in the assumptions.
Since in our formulation, depth positivity is enforced using the normal flow, which in-turn makes minimal assumptions about the scene structure, our formulation generalizes to novel scenes with remarkable accuracy (Sec. \ref{subsec:generalization}). 

Let us mathematically define the constraints. Denoting the magnitude of the normal flow (a scalar) at  pixel $\mathbf{x}$ as $n_\mathbf{x}$  and
the direction of the image gradient as $\mathbf{g_x}$ (a unit vector), we have
\begin{equation}\label{eq:un}
{n}_{\mathbf{x}} = \Vert\mathfrak{n}\Vert_2 =  \frac{1}{Z_{\mathbf{x}}}(\mathbf{g}_{\mathbf{x}}\cdot A)\mathbf{V} + (\mathbf{g}_{\mathbf{x}}\cdot B)\mathbf{\Omega},
\end{equation} 

where, $\mathbf{V}$  denotes the constant translational velocity and $\mathbf{\Omega}$ the rotational velocity $\mathbf{\Omega}$ in a small time instant, and $Z_{\mathbf{x}}$ is the depth of the point under consideration.
Intuitively, $\mathbf{V}, \mathbf{\Omega}$ warp the flow field and $Z_{\mathbf{x}}$ scales the flow field, and the matrices $A$ and $B$ determine how the motion flow field is projected onto the image plane due to translational and rotational velocities respectively and are given by

\begin{equation}\label{Ax}
A = \begin{bmatrix}
-1 & 0 & x\\
0 & -1 & y
\end{bmatrix}
\end{equation}

\begin{equation}\label{Bx}
B = \begin{bmatrix}
xy & -(x^2+1) & y\\
(y^2+1) & -xy & -x
\end{bmatrix}
\end{equation}

Let us consider in eq. (\ref{eq:un}) the two components of normal flow: the translational component, which depends on depth, and the rotational component, which is independent of depth. If we subtract the rotational component from both sides, we obtain
\begin{equation}\label{eq:un2}
{n}_{\mathbf{x}} - (\mathbf{g}_{\mathbf{x}}\cdot B)\mathbf{\Omega} = \frac{1}{Z_{\mathbf{x}}}(\mathbf{g}_{\mathbf{x}}\cdot A)\mathbf{V}.
\end{equation} 
We  can enforce that the left hand side (the derotated normal flow) and the right hand side (the translational component) must have same sign. Since the depth ($Z_{\mathbf{x}}$) is positive, the following product, which we denote as $\rho_\mathbf{x}(\mathbf{V},\mathbf{\Omega})$, is positive, i.e.,

\begin{equation}
\label{Zx}
\rho_\mathbf{x}(\mathbf{V},\mathbf{\Omega}) =  
\left(\left(\mathbf{g_x}\cdot A\right)\mathbf{V}\right) \cdot 
\left( {n_\mathbf{x}} - \left( \mathbf{g_x}\cdot B \right)\mathbf{\Omega} \right) > 0
\end{equation}

To arrive at an objective function to be used in an optimization, we can model the cheirality constraint by passing $\rho_\mathbf{x}$ through  a smooth function, such as the   ReLU function \cite{barranco2021joint}. Since, the deep learning pipeline requires the function to be twice differentiable, we choose the GELU function, a smooth approximation of the ReLu function. Denoting the negative GELU function as  $\mathcal{R}$, and denoting the  average over all $\mathbf{x}$ values as $\mathbb{E}$, we then obtain the following minimization, for the estimation of relative camera pose:
\begin{equation}
\label{eq:Lc}
\argmin_{\{\mathbf{V}, \mathbf{\Omega}\}} \mathbb{E}\left( \mathcal{R}(\rho_\mathbf{x}(\mathbf{V},\mathbf{\Omega})\right)
\end{equation}





In the stage of  re-estimating motion, we simply use constraint (\ref{eq:Lc}) in an optimization implemented by a robust Quasi-Newton optimization algorithm.  We solve the optimization sequentially, in one step for $V$ and in the other for $\Omega$, because $\rho_\mathbf{x}(\mathbf{V},\mathbf{\Omega})$ is bilinear in these parameters. The initial estimate comes from the PoseNet estimate in Sec.~\ref{subsec:coarse_posenet}.
In our implementation we use the L-BFGS algorithm \cite{zhu1994bfgs}.
These steps (Sec. \ref{subsec:coarse_posenet} and \ref{subsec:cheirality}) are performed in the forward pass in the network.



\subsection{Self-Supervised Refinement}
\label{subsec:implicit}


Let us denote the coarse pose obtained from our \textit{PoseNet} as $\mathbf{\widetilde{P}}_c =  \begin{bmatrix} \mathbf{\widetilde{V}}_c & \mathbf{\widetilde{\Omega}}_c\end{bmatrix}$. This is further refined by the Cheirality layer and we denote this refined pose as $\mathbf{\widetilde{P}}_r = \begin{bmatrix} \mathbf{\widetilde{V}}_r & \mathbf{\widetilde{\Omega}}_r\end{bmatrix} $.
Now we use  $\mathbf{\widetilde{P}}_r$ to refine our \textit{PoseNet}'s coarse pose to obtain a more accurate prediction of pose. 

The  final self-adaptive pose estimation  is performed as a bi-level minimization in the network \cite{gould2021deep}, in which an upper-level problem is solved subject to constraints imposed by a lower-level problem and is formally defined next.



\begin{equation}
\label{eq:refloss}
    \begin{aligned}[t] &\argmin_{\mathbf{\widetilde{P}}_c} 
     \mathbb{E}\left({n_\mathbf{x}} - \mathbf{g_x} \cdot \left(\left(\frac{{n_\mathbf{x}} - (\mathbf{g_x} \cdot B) \widetilde{\mathbf{\Omega}}_r}{(\mathbf{g_x} \cdot A) \widetilde{\mathbf{V}}_r}\right) A\widetilde{\mathbf{V}}_c -  B \widetilde{\mathbf{\Omega}}_c \right) \right)      \\
    & \text{subject to} \quad
    \argmin_{\mathbf{\widetilde{P}}_r} \mathbb{E}\left( \mathcal{R}(\rho_\mathbf{x}(\widetilde{\mathbf{V}}_c,\widetilde{\mathbf{\Omega}}_c))\right)
     \end{aligned}
\end{equation}

The lower-level problem (second row) of Eq. (\ref{eq:refloss}) enforces the cheirality constraint 
to obtain   pose $\mathbf{\widetilde{P}}_r$, which then is used to compute the normal flow error in the upper-level loss function (first row). The upper-level loss enforces the consistency between the normal flow from  \textit{NFlowNet} and that computed using
the motion parameters $\mathbf{\widetilde{P}}_c$
with the implicit depth term expressed by the motion parameters $\mathbf{\widetilde{P}}_r$.


In practice, we back-propagate through the upper-level to refine poses $\mathbf{\widetilde{P}_c}$ using supervision from $\mathbf{\widetilde{P}_r}$ through  the lower-level.
Using implicit differentiation, all that is computed from the lower layer is the gradient, and this step is agnostic
to the optimizer used. Specifically, we derive  $\frac{\partial \mathbf{\widetilde{P}}_r}{\partial \mathbf{\widetilde{P}}_c}$, which is computed from the product of second order derivatives. (The interested reader is referred to \cite{gould2021deep}, eq. (15) for  details.)
It is important to note that we rely on the generalizability of \textit{NFlowNet}, hence it is not fine-tuned.




     


\section{Experiments}
\label{sec:experiments}

\subsection{Implementation Details}
\subsubsection{Datasets} We use eight environments from the TartanAir \cite{tartanair2020iros}  dataset (\texttt{amusement}, \texttt{oldtown}, \texttt{neighborhood}, \texttt{soulcity}, \texttt{japanesealley}, \texttt{office}, \texttt{office2}, \texttt{seasidetown}) for training and two environments (\texttt{abandonedfactory} and \texttt{hospital}) for testing our \textit{NFlownet} network. For odometry evaluation, we use the Tartan challenge test data \cite{tartanair2020iros}. We also conduct extensive experiments on the KITTI Odometry \cite{kitti} and the TUM-RGBD \cite{sturm12iros} datasets to evaluate the robustness and generalization performance of our proposed system.  
\subsubsection{Networks and Optimization Layer} 
For the \textit{NFlowNet} we use an encoder-decoder architecture based on  EVPropNet \cite{sanket2021evpropnet} to directly regress sparse normal flow. The encoder contains residual blocks with convolutional layers and the decoder  contains residual blocks with transpose convolutional layers. We choose the number of residual and transposed residual blocks as 2 and the expansion factor (factor by which the number of neurons are increased after every block) as 2. We backpropagate the gradients using a mean squared loss computed between groundtruth and predicted normal flow as given in Eq. \ref{eqn:nfloss}.  We used the Adam optimizer  to train our network with a learning rate of $10^{-4}$ and  batch size of 8 for 400 epochs. 

Our \textit{PoseNet} architecture is inspired from \cite{wang2017deepvo} and uses the VGG-16 encoder for the CNN stage \cite{dosovitskiy2015flownet} and two LSTM layers each with 250 hidden units for the recurrent layer-stage. We initially train this model with a subset of the TartanAir data for 30 epochs to obtain a coarse  estimate to initialise the Cheirality Layer. We use the Adam optimizer and set a fixed learning rate of 10$^{-5}$. We consider sequences of six consecutive image frames and a batch size of eight while training. During test time, we use only two consecutive image frames to estimate the relative camera pose between images $I_{t}$ and $I_{t+1}$.   

For optimization layer, We use the L-BFGS \cite{lbfgs} solver. The line search function was set to strong Wolfe \cite{strongwolfe}, the number of iterations were set to 100 and the gradient norms were clipped to 100. We initialised the optimizer with coarse predictions provided by our \textit{PoseNet}. Our overall system is implemented in Python 3.7 and PyTorch 1.9.

\subsubsection{Training and Testing Procedure} The whole training schedule consists of three stages. First, we only train \textit{NFlowNet} and \textit{PoseNet} in a supervised manner using the training strategies mentioned above. Then, we freeze \textit{NFlowNet} and jointly train the \textit{PoseNet} with the cheirality layer in an self-supervised fashion via the refinement loss. The self supervised training is carried out for 120 epochs using four Nvidia P6000 GPUs. For cheirality layer, the stopping criteria is when the objective function is below $10^{-20}$ or the number of iterations exceeds 300.

During testing, our final pose predictions are obtained by passing  \textit{PoseNet} priors through the cheirality layer along with the predictions from \textit{NFlowNet}. Due to our self-supervised refinement training, the prior \textit{PoseNet} estimates help in faster convergence of the cheirality optimization process (takes less than 5 iterations). We will release codes for inferencing \textit{NFlowNet} and our differentiable chierality layer upon the acceptance of the paper.

\subsubsection{Evaluation Metrics} 
To evaluate our \textit{NFlowNet} and to compare against other optical flow networks, we project the optical flow obtained from the state-of-the-art approaches on the gradient direction, and we measure the pixel error. As normal flow is defined  only along the gradient direction, we utilize the Projection Endpoint Error (PEE), an analog to the Average Endpoint Error (AEE) error metric proposed in \cite{Hordijk2018VerticalLF}. The PEE between the projected optical flow ($\widetilde{\mathbf{u}}$) and the groundtruth ($\hat{\mathbf{\mathfrak{n}}}$) is defined as:

\begin{equation}
     \text{PEE}= \Big\Vert \hat{\mathbf{\mathfrak{n}}}- \frac{ \nabla I}{||\nabla I||_2^2} \cdot \widetilde{\mathbf{u}} \Big\Vert
\end{equation}

To evaluate the regressed relative poses from our model, we use two metrics:

\textbf{Absolute Trajectory Error:} \cite{metrics} evaluates the global consistency by comparing the absolute poses between the estimated and
the ground truth trajectory. As the predicted and groundtruth trajectories can be
specified in arbitrary coordinate frames, they are aligned using \cite{Horn:87}, which finds the 
rigid-body transformation $S$. The error matrix between poses, $E_t$  is  defined as:

\begin{equation}
    \label{eqn:ATE}
    E_{t} =  Q_t^{-1} S P_t
\end{equation}

Here, $P_t \in SE(3)$ is predicted pose at time $t$ from the sequence of the spatial poses, $Q_t \in SE(3)$ is pose at time $t$  from the ground truth trajectory, and $E_t$ is the error matrix between these poses. Then, ATE is defined as the root mean square error of the error matrices:
\begin{equation}
ATE =    \left (  \frac{1}{n} \sum _{i=t}^n \left \| \text{trans}(E_t) \right \|^2 \right )^{\frac{1}{2}}
\end{equation}

\textbf{Relative Pose Error:} measures the local trajectory accuracy over a fixed time interval $\Delta t$. From a sequence of $n$ poses, we obtain $m=n-\Delta t$ relative pose error matrices $F_t^{\Delta t}= ( Q_t^{-1} Q_{t + \Delta t})^{-1} ( P_t^{-1} P_{t + \Delta t})$
\begin{equation}
    t_{rel} = RPE_{\text{trans}}^{t, \Delta t}  = \left (  \frac{1}{m} \sum _{t=1}^m \left \| trans(F_t^{\Delta t}) \right \|^2 \right )^{\frac{1}{2}}
\end{equation}
\begin{equation}
r_{rel} =  RPE_{rot}^{t, \Delta t}  = \frac{1}{m} \sum _{t=1}^m \angle \text{rot}(F_t^{\Delta t}), 
\end{equation}
where $\text{trans}(\cdot)$ and $\text{rot}(\cdot)$ denote the translation and rotation components of the pose matrix. $\angle R  = arccos\left(\frac{\text{tr}(R)-1}{2}\right)$ with  $R$ denoting the rotation matrix.

\subsection{Analysis of Experimental Results} 
\subsubsection{Accuracy of Normal Flow}

In the first case study, we  quantitatively evaluate \textit{NFlowNet}. We compare our network with optical flow methods of various flavours: (a) supervised (PWC-Net \cite{pwc-net}, LiteFlowNet \cite{liteflownet}), (b) self-supervised (SelFlow \cite{selflow}). 


In Table \ref{tab:nfcomparetable}, we present a quantitative evaluation of normal flow. We trained our \textit{NFlowNet} and fine-tuned optical flow networks with TartanAir (8 environments). We demonstrate the PEE error on the first four sequences of the environments \texttt{abandonedfactory} and \texttt{hospital}. \textit{NFlowNet} performs better than the optical flow networks by up to 6$\times$. By  learning normal flow, we constrain the network to focus on prominent features (edges, textures) rather than dense correspondences in textureless regions. 
Through this ``attention-like'' mechanism, \textit{NFlowNet} performs better than its optical flow counterparts with upto 46$\times$ smaller model size.


\begin{table}[h!]
    \centering
     \caption {PEE (Projection Endpoint Error) $\downarrow$ of different state-of-the-art methods as compared to of our \textit{NFlowNet}.}
    \resizebox{\columnwidth}{!}{
        \label{tab:nfcomparetable}
    \begin{tabular}{llllllllllll}
    \toprule
         \multirow{2}{*}{Method}

         & \multicolumn{4}{c}{\texttt{abandonfactory}}
         & \multicolumn{4}{c}{\texttt{hospital}} & Num. \\
        
         & 000 & 001 & 002 & 003 & 000 & 001 & 002 & 003 &  param.(M)  $\downarrow$\\
         \hline
         
         LiteFlownet \cite{liteflownet} & 2.56 & 1.82 & 1.93 & 2.15 & 3.17 & 2.68 & 2.45 & 1.93 & 5.37\\
         PWC-Net \cite{pwc-net} & 1.23 & 0.95 & 1.64 & 1.48 & 2.35 & 2.92 & 2.28 & 1.47 &  8.75\\
         UnFlow \cite{unflow} & 1.35 & 1.15 & 1.75 & 1.56 & 2.21 & 1.76 & 1.83 & 1.07 & 126.90 \\
         SelFlow \cite{selflow} & \textbf{0.67} & 0.73 & \textbf{0.52} & 0.64 & 1.91 & 0.51 & 0.73 & \textbf{0.65} & 5.11 \\
         \hline
         \textit{NFlowNet} (Ours) & 0.72 & \textbf{0.54} & 0.57 & \textbf{0.63} &\textbf{ 0.82} & \textbf{0.44} & \textbf{0.57} & 0.71 & \textbf{2.72} \\
         \bottomrule
    \end{tabular}}\\
\end{table}

\subsubsection{Comparison of Pose Estimation}
\label{subsec:generalization}

In this section we compare  our \textit{DiffPoseNet} framework  with various state-of-the-art camera pose estimation approaches. These approaches can be broadly be classified 
into: (a) pure deep learning models, (b) pure geometric constraint based models and (c) hybrid deep learning models that incorporate some form of geometric constraints in the learning pipeline.

We present the Absolute Trajectory Error (ATE) of our model on the TartanAir challenge data in the \texttt{MH000-007} sequences and compare it to the results of TartanVO and ORB-SLAM in Table \ref{tab:tartan}. We outperform both methods  in this experiment by up to 3.4$\times$.  

We present the ATE in Table \ref{tab:tum_compare} for the  TUM-RGBD sequences (\texttt{360}, \texttt{desk}, \texttt{desk2},  \texttt{rpy}, \texttt{xyz}). We achieve competitive results, because TUM RGBD is difficult for monocular vision methods due to rolling shutter, motion blur and large rotation. This is where pure geometric constraints show a massive advantage. We believe our work can  further be improved by compensating for rolling shutter in the differential layer and we see this as a potential future research direction. {It is important to stress that our method was trained on the  TartanAir dataset and tested directly on other datasets to highlight cross-dataset generalization without any fine-tuning or re-training.}

\begin{figure*}
\centering
\resizebox{\textwidth}{!}{
\begin{tabular}{cccc}
\subcaptionbox{\texttt{Seq. 05}\label{1}}{\includegraphics[width = 0.25\textwidth]{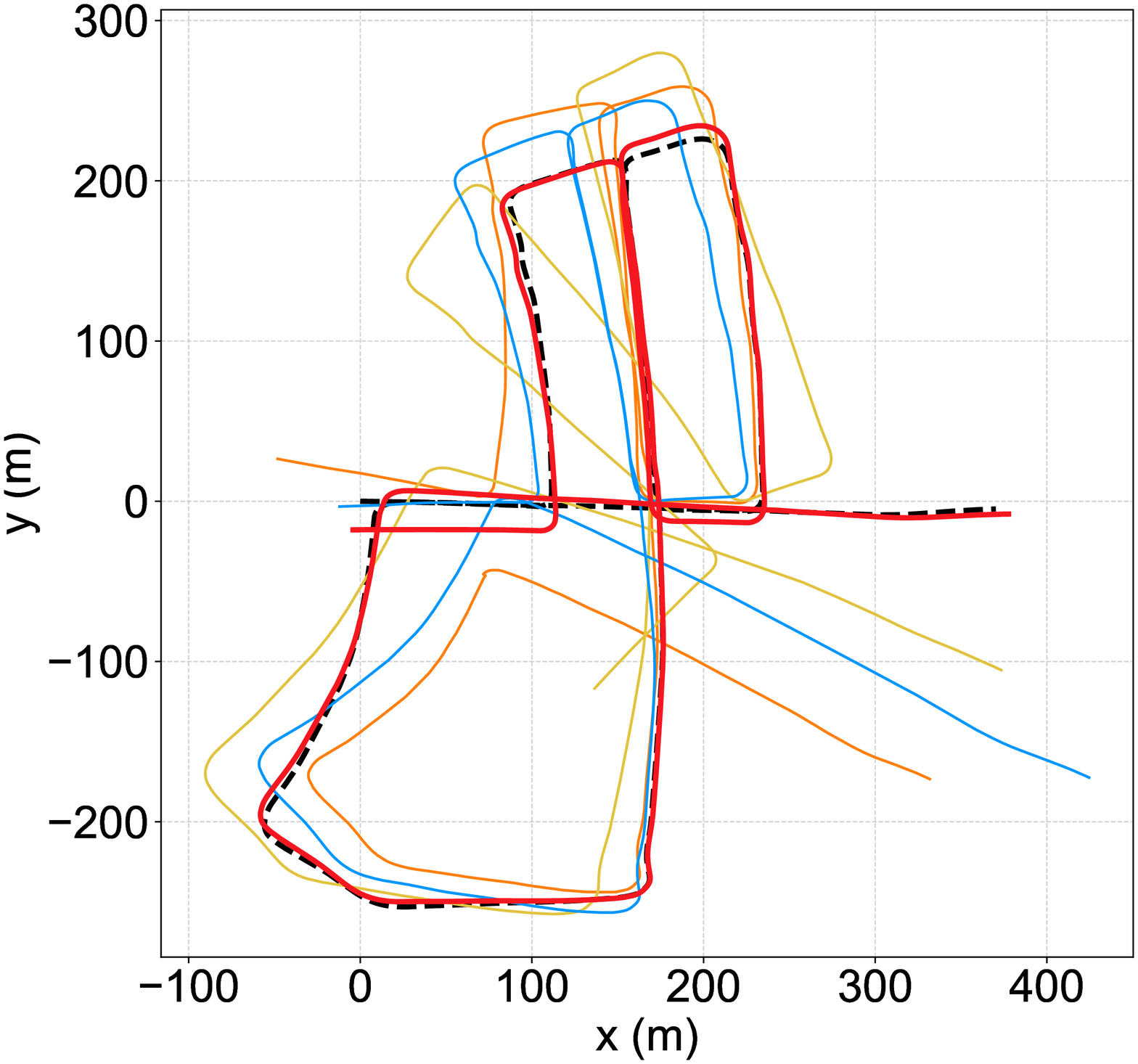}}

&
\subcaptionbox{\texttt{Seq. 07}\label{2}}{\includegraphics[width = 0.25\textwidth]{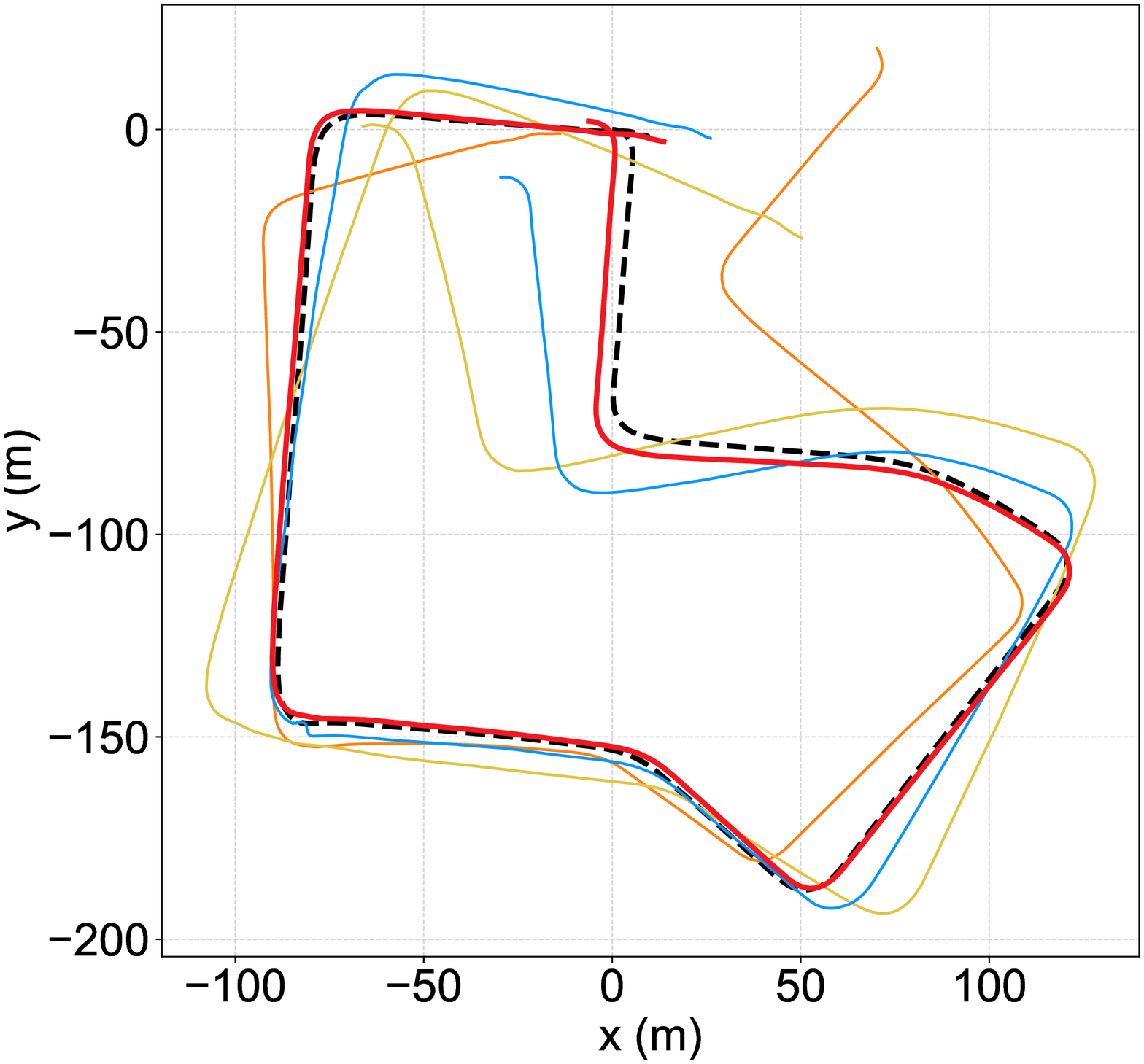}}
&
 \subcaptionbox{\texttt{Seq. 09}\label{1}}{\includegraphics[width = 0.25\textwidth]{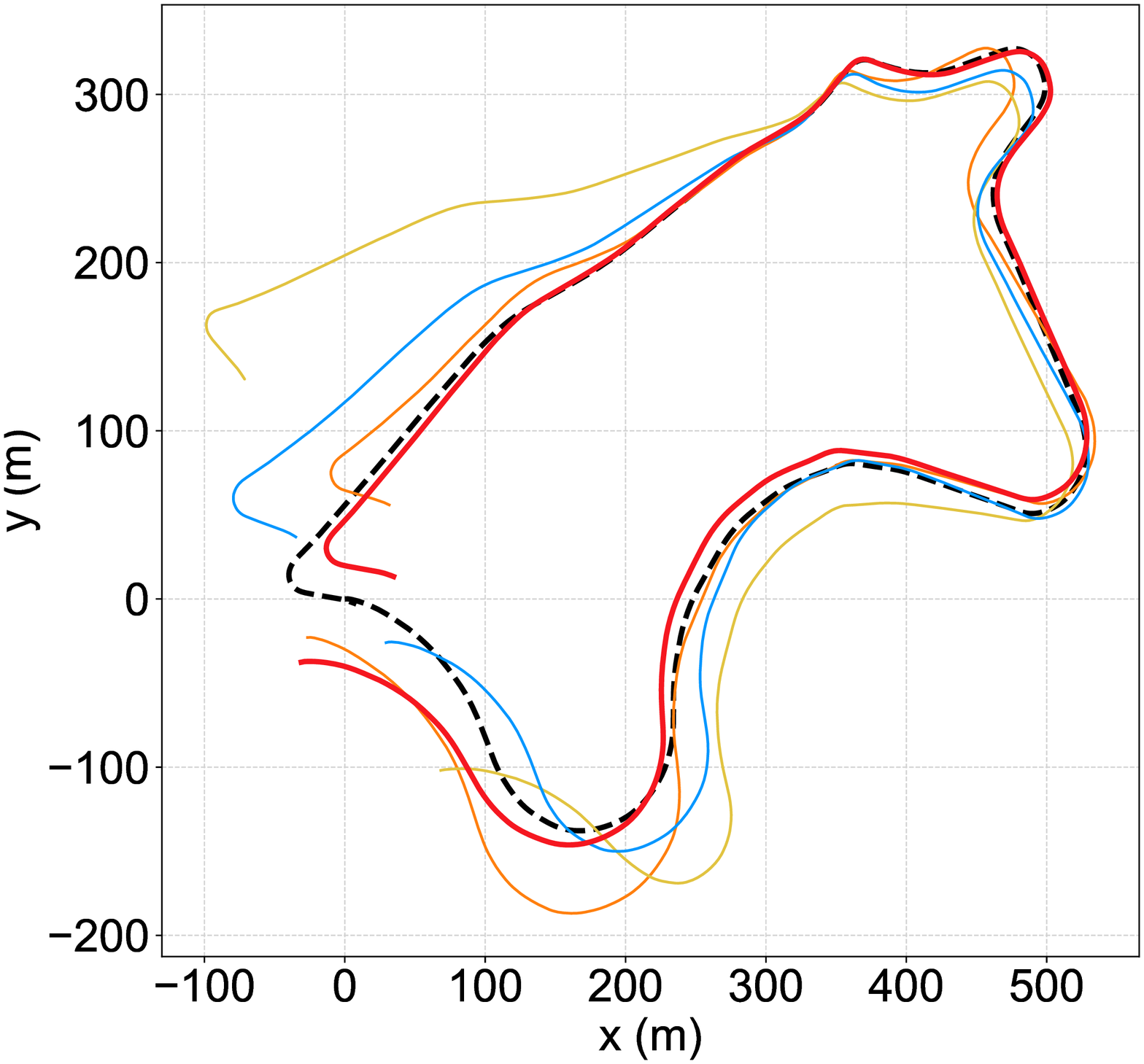}}
 &
\subcaptionbox{\texttt{Seq. 10}\label{2}}{\includegraphics[width = 0.25\textwidth]{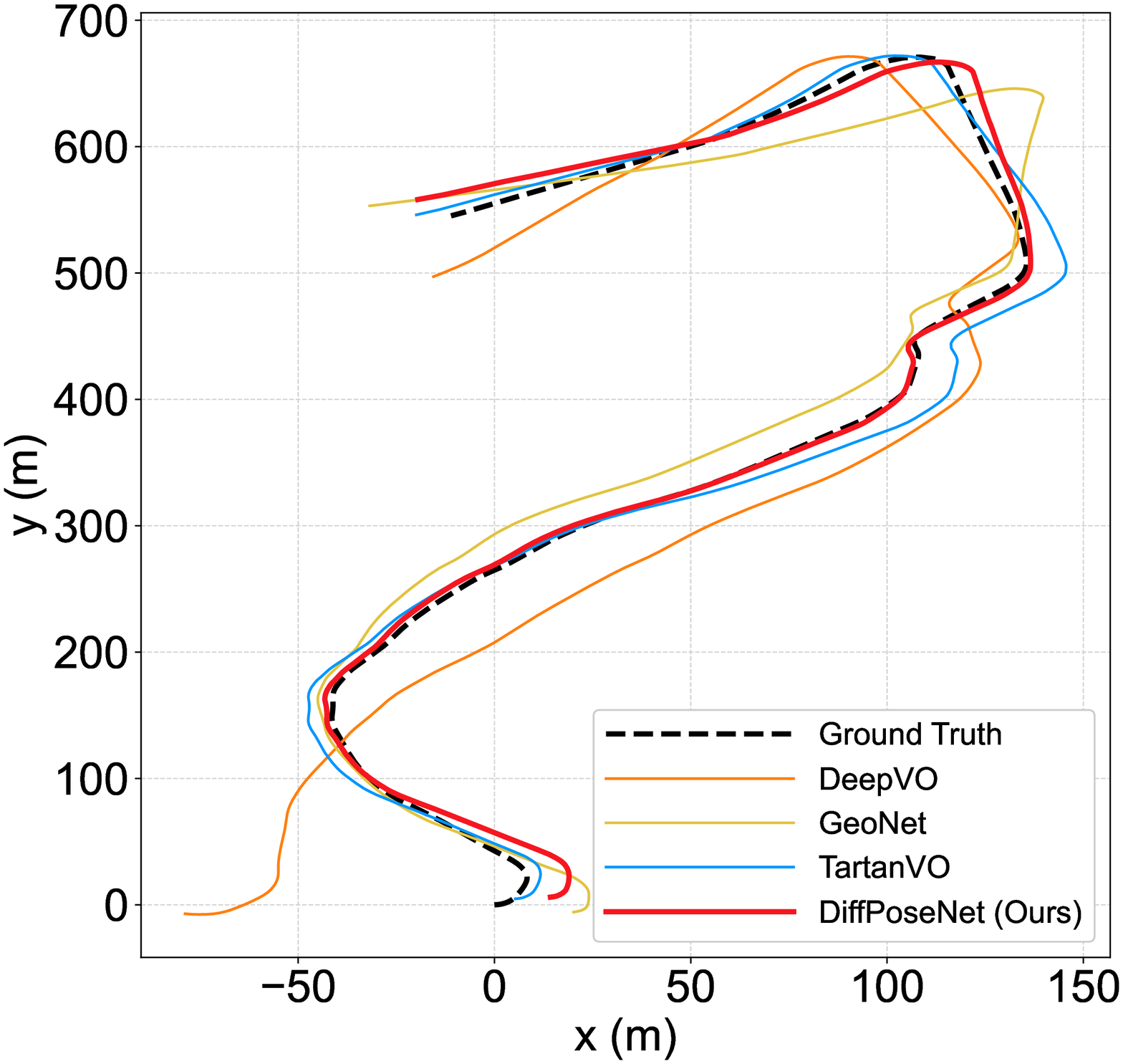}}
\end{tabular}}
\caption{Qualitative comparison of trajectories between our \textit{DiffPoseNet} and other state-of-the-art approaches on the KITTI dataset.}
\label{fig:kittiodom}
\end{figure*}

We also present the relative pose errors, specifically, the average translational RMSE drift ($t_{rel}$ in \% ) and average rotational RMSE drift ($r_{rel}$ in $^\circ$/100m) for comparison in the KITTI dataset sequences \texttt{06}, \texttt{07}, \texttt{09} and \texttt{10} (See Fig. \ref{fig:kittiodom}). The error metrics are computed on a trajectory of length of 100–800 m. 
In Table \ref{tab:kitti} we compare our model with (a) pure deep learning models (TartanVO\cite{wang2020tartanvo}, GeoNet\cite{yin2018geonet}, UnDeepVO\cite{li2018undeepvo}, DeepVO \cite{wang2017deepvo}, Wang et al. \cite{Wang-2019-118682}), (b) pure geometric constraint based approaches (ORB-SLAM\cite{orb-slam}, VISO2-M\cite{viso2-m}) and with (c) our \textit{DiffPoseNet}, BiLevelOpt\cite{jiang2020joint}.
 The test sequences were selected such that they do not overlap with the training sets in the deep learning models used for comparison. Note that, similar to TartanVO, our training is performed only on TartanAir and do not perform any fine-tuning or re-training on KITTI dataset. Regardless, we perform competitive to other approaches that are trained/fine-tuned on similar data, thus demonstrating our cross-dataset generalization.  

\begin{figure}
\includegraphics[width =\columnwidth]{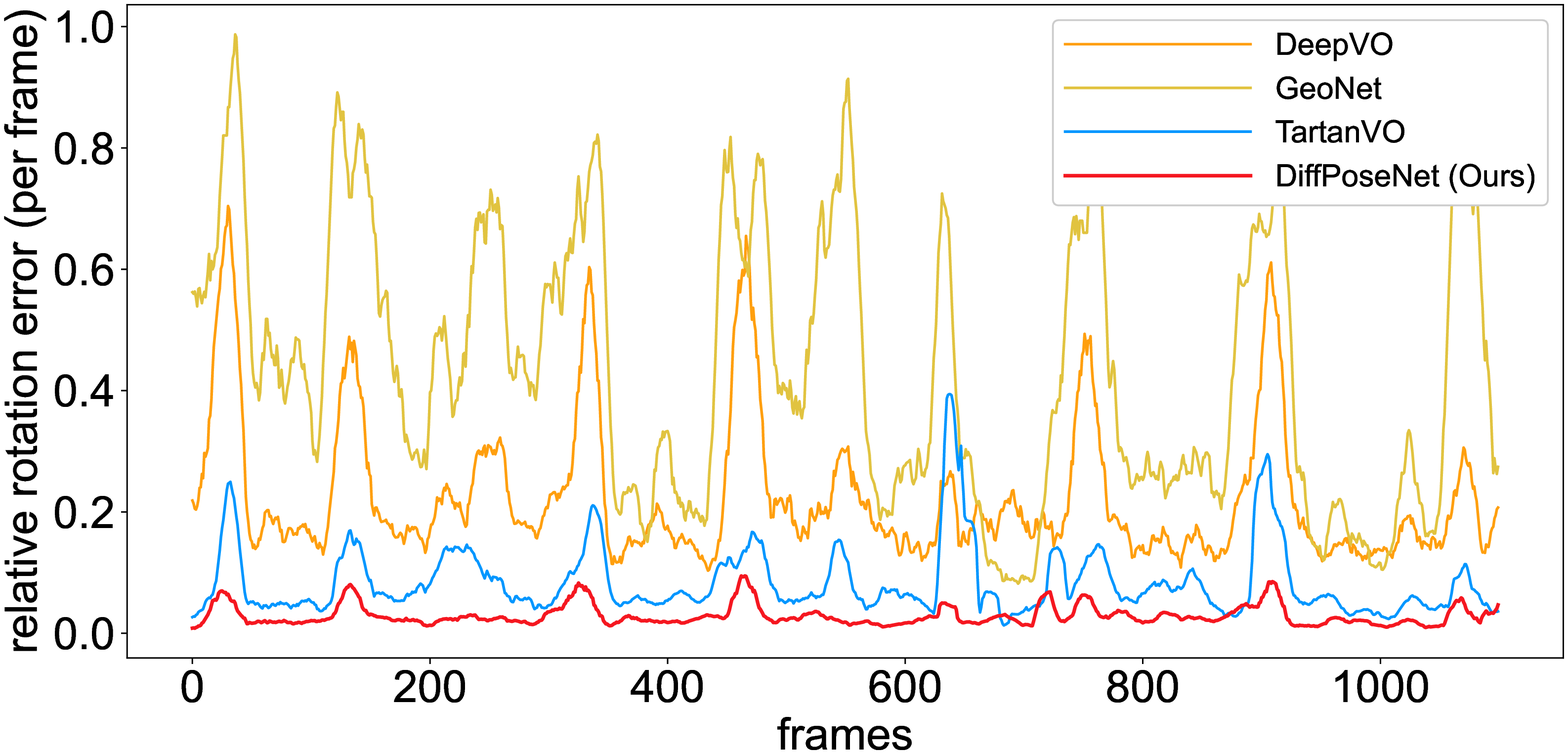}
\caption{Comparison between our model and pure learning based VO on relative rotation error (in $^{\circ}$/frame) from \texttt{KITTI - 07}.} 
\label{fig:rotations}
\end{figure}

We infer from the above results (also see Fig. \ref{fig:rotations}) that deep learning models usually outperform classical geometric constraint based methods in translation errors ($t_{rel}$), which can be attributed to the scale drift issue. This sometimes might be solved by performing expensive global bundle adjustment and loop closure. However, we are not using a loop closing procedure in the  models in our experiments. Models with differentiable optimizer layers, like \textit{DiffPoseNet} (Ours) and BilevelOpt \cite{jiang2020joint} achieve the best of both worlds, with lower relative rotation errors competitive with geometric methods like ORB-SLAM.



    

\begin{table}[t!]
    \centering
     \caption {ATE (m) $\downarrow$ on the \texttt{MH} sequences of TartanAir \cite{tartanair2020iros} dataset.}
    \resizebox{\columnwidth}{!}{
        \label{tab:tartan}
    \begin{tabular}{lllllllllll}
    \toprule
    
        Methods & \texttt{000} & \texttt{001} & \texttt{002} & \texttt{003} & \texttt{004} & \texttt{005} & \texttt{006} & \texttt{007}\\
         
        
         
        \hline
        ORB-SLAM \cite{orb-slam} & \textbf{1.30} & \textbf{0.04} & 2.37 & 2.45 & - & - & 21.47 & 2.73 \\
        TartanVO \cite{wang2020tartanvo} & 4.88 & 0.26 & 2.00 & 0.94 & 1.07 & 3.19 & \textbf{1.00} & 2.04 \\
         \hline
         Ours & 2.56 & 0.31 & \textbf{1.57} & \textbf{0.72} & \textbf{0.82} & \textbf{1.83} & 1.32 & \textbf{1.24}\\
         \bottomrule
    \end{tabular}}
\end{table}

\begin{table}[t!]
    \centering
     \caption {ATE (m) $\downarrow$ on the TUM-RGBD \cite{sturm12iros} benchmark.}
    \resizebox{\columnwidth}{!}{
        \label{tab:tum_compare}
    \begin{tabular}{llllll}
    \toprule
    
        Methods & \texttt{360} & \texttt{desk} & \texttt{desk2} & \texttt{rpy} & \texttt{xyz}\\
         
        
         
        \hline
        ORB-SLAM2 \cite{orb-slam2} & - & \textbf{0.016} & 0.078 & - & \textbf{0.004}  \\
       \hline
        DeepTAM \cite{deeptam} & \textbf{0.116} & 0.078 & 0.055 & 0.052 & 0.054 \\
        TartanVO \cite{wang2020tartanvo} & 0.178 & 0.125 & 0.122 & \textbf{0.049} & 0.062 \\
         \hline
         Ours & 0.121 & 0.101 & \textbf{0.053} & 0.056 & 0.048 \\
         \bottomrule
    \end{tabular}}
\end{table}

\begin{table}[t!]
    \centering
     \caption {Relative Pose Error ($t_{rel}$ and $r_{rel}$) $\downarrow$ results of various Pose estimation methods on KITTI \cite{kitti} dataset. Note that, \textbf{bold} represents the best result and \underline{underline} represents the second best.}
    \resizebox{\columnwidth}{!}{
        \label{tab:kitti}
    \begin{tabular}{lllllllll}
    \toprule
         \multirow{2}{*}{Method}
         
        
         & \multicolumn{2}{c}{\texttt{06}}
         & \multicolumn{2}{c}{\texttt{07}}
         & \multicolumn{2}{c}{\texttt{09}} 
         & \multicolumn{2}{c}{\texttt{10}}\\ 
         \cline{2-9} \\ [-5pt] 
         & $t_{rel}$ & $r_{rel}$ &$t_{rel}$ & $r_{rel}$ & $t_{rel}$ & $r_{rel}$ & $t_{rel}$ & $r_{rel}$  \\
         \hline
         
           DeepVO\cite{wang2017deepvo} & 5.42 & 5.82 & \underline{3.91} & 4.60 & - & - & 8.11 & 8.83\\
          Wang et al.\cite{Wang-2019-118682} & - & - & - & - & 8.04 & 1.51 & 6.23 & 0.97\\
          UnDeepVO \cite{li2018undeepvo} & 6.20 & 1.98 & \textbf{3.15} & 2.48 & - & - & 10.63 & 4.65 \\
          GeoNet \cite{yin2018geonet} & 9.28 & 4.34 & 8.27 & 5.93 & 26.93 & 9.54 & 20.73 & 9.04 \\
          TartanVO \cite{wang2020tartanvo} & \underline{4.72} & 2.95 & 4.32 & 3.41 & 6.03 & 3.11 & 6.89 & 2.73  \\
          BiLevelOpt \cite{jiang2020joint} & - & - & - & - & 4.36 & 0.69 & 4.04 & 1.37  \\
         \hline
         ORB-SLAM \cite{orb-slam} & 18.68 & \textbf{0.26} & 10.96 & \textbf{0.37} & 15.3 & \textbf{0.26} & \textbf{3.71} & \textbf{0.3}  \\
         VISO2-M \cite{viso2-m} & 7.34 & 6.14 & 23.61 & 19.11 & \underline{4.04} & 1.43 & 25.2 & 3.84  \\
         \hline
         Ours & \textbf{2.94} & \underline{1.76} & 4.06 & \underline{2.35} & \textbf{4.02} & \underline{0.51} & \underline{3.95} & \underline{1.23} \\
         \bottomrule
    \end{tabular}}\\
\end{table}

\subsubsection{Robustness of Pose Estimation}

\begin{figure}

\includegraphics[width =\columnwidth]{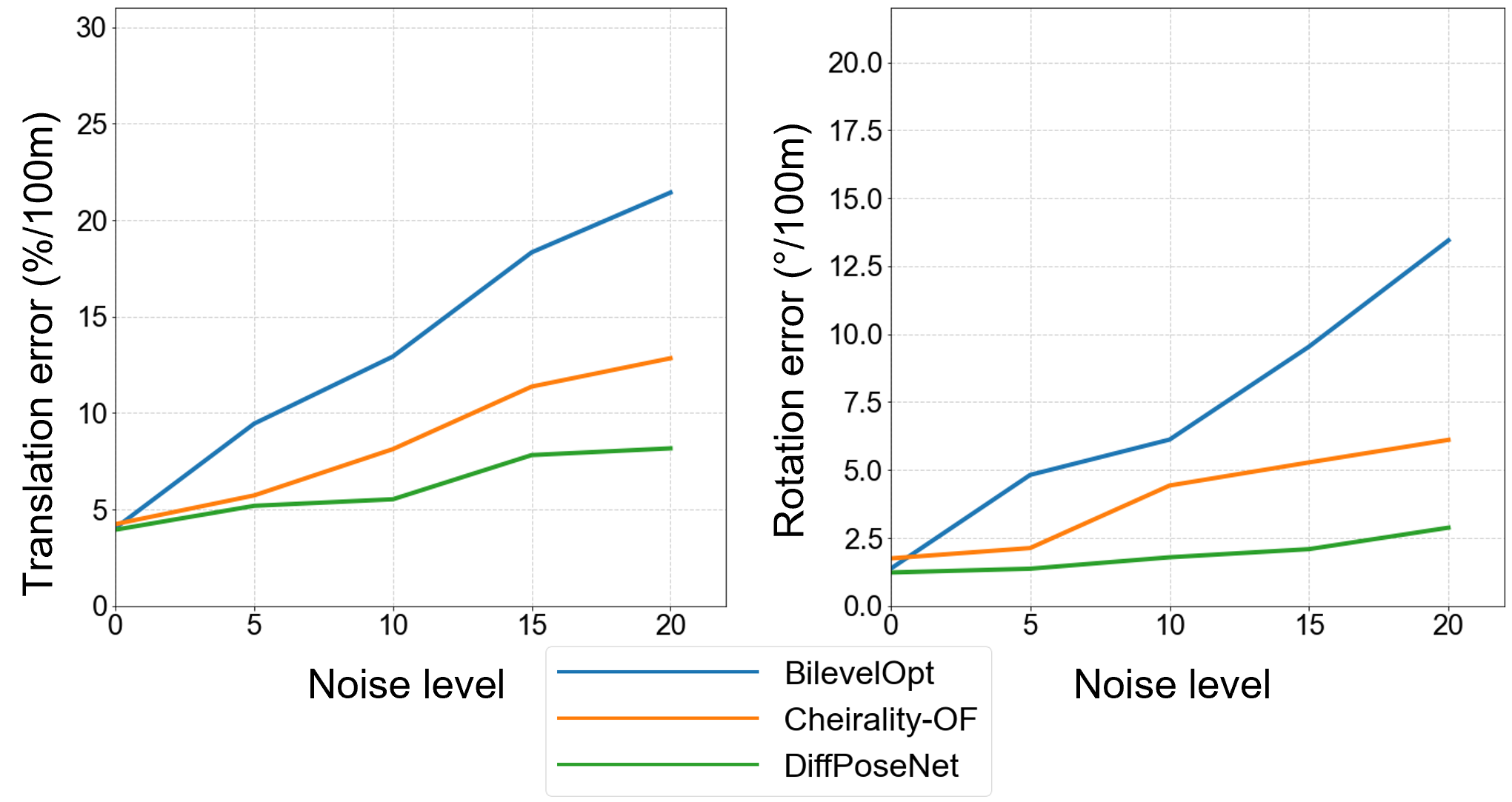}
\caption {Robustness evaluation of normal flow representation for pose estimation on \texttt{KITTI-10}.}
\label{fig:robustness}
\end{figure}

For most camera pose estimation approaches, the performance of the algorithm is also governed by external factors such as lighting, weather, and sensor noise. These external factors often lead to errors in motion fields and cause pose estimation to fail or diverge. In this case study we present a robustness analysis 
by injecting noise into the motion fields. 

We study the robustness of the normal flow based cheirality layer against the epipolar layer \cite{jiang2020joint}.
We artificially inject errors in the  normal flow and optical flow and evaluate our framework's performance under these conditions. The error is modelled as  additive uniform noise $\mathcal{U}\left(\epsilon\right)$, where $[-\epsilon,\epsilon]$ is the bound on noise. We induce this noise to both normal flow and optical flow only on the gradient direction where normal flow is well-defined to make the comparison fair. 

Fig. \ref{fig:robustness} shows the relative pose errors, $t_{rel}$ and $r_{rel}$, over  $\epsilon$ values  $\{0,5,10,15, 20\}\%$. Here, BiLevelOpt refers to the pose estimation using the epipolar constraint layer \cite{jiang2020joint} with optical flow as input. ``Cheirality-OF" refers to the pose estimation via the cheirality layer using normal flow obtained by projecting SelFlow optical flow predictions on the gradient directions (we choose SelFlow as it has the best performance among all optical flow methods used in this paper). Observe that, our \textit{DiffPoseNet} is more resilient to noise and fails ``more gracefully'' compared to other methods. We owe this robustness to the lack of strong constraints used in our approach as compared to other methods that either rely on strong features or strong photometric consistency. We believe that carefully crafted optimization problems can lead to robust pose estimation neural networks that generalize well to novel datasets while being robust to noise, a capability rarely seen  in most state-of-the-art methods \cite{9356338, 9010809}. 






\section{Conclusion}
In this work, we combined the best of both worlds from classical direct camera pose estimation approaches and deep learning taking advantage of  differentiable programming concepts. Specifically, we addressed the problem of estimating relative camera pose using a sequence of images. To achieve this, we introduced the \textit{DiffPoseNet} framework. As part of this framework, we introduced a normal flow network called \textit{NFlowNet}, that predicts  accurate motion fields under challenging scenarios and is more resilient to noise and bias. Furthermore, we proposed a differentiable cheirality layer that when coupled with \textit{NFlowNet} can estimate robust and accurate relative camera poses. 

A comprehensive qualitative and quantitative evaluation was provided on the challenging datasets: TartanAir, TUM-RGBD and KITTI. We demonstrated the efficacy, accuracy and robustness of our method under noisy scenarios and cross-dataset generalization without any fine-tuning and/or re-training . Our approach outperforms the previous state-of-the-art approach. Particularly, NFlowNet can output accurate up to 6$\times$ motion fields with up to 46$\times$ smaller model size. We believe this will open a new direction for  camera pose estimation problems. 

\label{sec:conclusion}

{\small
\bibliographystyle{ieee_fullname}
\bibliography{egbib}
}

\end{document}